\documentclass[journal]{IEEEtran}

\usepackage{cite}
\usepackage{graphicx}
\graphicspath{ {Figures/} }
\usepackage{caption}
\usepackage{subcaption}
\usepackage{epstopdf}

\usepackage{array}
\usepackage{amsmath}
\usepackage{amssymb}
\usepackage{bm}

\usepackage{hyperref}

\usepackage[ruled,lined,commentsnumbered]{algorithm2e}

\def\b{{\mathbf b}}

\def\x{{\mathbf x}}
\def\y{{\mathbf y}}
\def\A{{\mathbf A}}
\def\B{{\mathbf B}}
\def\X{{\mathbf X}}
\def\Y{{\mathbf Y}}

\begin{document}

\title{Metalearning: Sparse Variable-Structure Automata}

\author{Pedram~Fekri, 
        Ali~Akbar~Safavi, 
        Mehrdad~Hosseini~Zadeh, 
        and~Peyman~Setoodeh
\thanks{P. Fekri is with the Department of Mechanical, Industrial, and Aerospace Engineering, Concordia University, Montreal, QC, Canada (e-mail: p\_fekri@encs.concordia.ca).}
\thanks{A. A. Safavi and P. Setoodeh are with the School of Electrical and Computer Engineering, Shiraz University, Shiraz,
Iran (e-mail: safavi@shirazu.ac.ir; psetoodeh@shirazu.ac.ir).}
\thanks{M. H. Zadeh is with the Department of Electrical and Computer Engineering, Kettering University, Flint, MI, USA (e-mail: mzadeh@kettering.edu).}
}

\maketitle

\begin{abstract}
Dimension of the encoder output (i.e., the code layer) in an autoencoder is a key hyper-parameter for representing the input data in a proper space. This dimension must be carefully selected in order to guarantee the desired reconstruction accuracy. Although overcomplete representation can address this dimension issue, the computational complexity will increase with dimension. Inspired by non-parametric methods, here, we propose a metalearning approach to increase the number of basis vectors used in dynamic sparse coding on the fly. An actor-critic algorithm is deployed to automatically choose an appropriate dimension for feature vectors regarding the required level of accuracy. The proposed method benefits from online dictionary learning and fast iterative shrinkage-thresholding algorithm (FISTA) as the optimizer in the inference phase. It aims at choosing the minimum number of bases for the overcomplete representation regarding the reconstruction error threshold. This method allows for online controlling of both the representation dimension and the reconstruction error in a dynamic framework.
\end{abstract}


\section{Introduction}
\label{intro}

Sparse coding represents inputs by generating a code vector~\cite{Olshausen1996, Olshausen1997}. The code includes multiple atoms that each of them determines which patterns would be involved in reconstructing the input. Consequently, similar inputs have codes with common active atoms. Sparse coding can be considered as a representation learning method, which attempts to extract robust features from input data in an over-complete manner. Sparsity can improve the robustness of the extracted features by omitting weak active atoms. Autoencoders can be built around sparse coding. An autoencoder is a generative unsupervised deep neural network that extracts robust features by adding a sparsity constraint term to its cost function. Using the trait of sparsity in an autoencoder causes to discard weak active hidden units for obtaining robust representations~\cite{Makhzani2014}.

The learning criterion in sparse coding is the Euclidean norm of the difference between input signals and their reconstructions (i.e. linear combinations of dictionary elements) \cite{Mairal2012}:
\begin{eqnarray} \label{eq:energy}
\mathcal{E}\left(\B,\x \right) = \frac{1}{2} \| \y-\B \x \|^2_2
\end{eqnarray}
where $\y \in \mathbb{R}^m$  is the input vector and $\B \in \mathbb{R}^{m \times n}$ is the dictionary matrix, whose columns form an over-complete basis. Learning over-complete representations in sparse coding means that the number of basis vectors (i.e. dictionary elements) is larger than the input dimension ($n>>m$). $\x \in \mathbb{R}^n$ denotes the corresponding coefficient vector. It is worth noting that the energy function in (\ref{eq:energy}) is under-determined. In order to seek robust features, a sparsity regularization term is added to the objective function in (\ref{eq:energy}). $\ell_1$-norm regularization is usually used for this purpose due to its sparsity property. Thus, the energy function for sparse coding is modified as follows~\cite{Mairal2012}:
\begin{eqnarray} \label{eq:energy2}
\mathcal{E}\left(\B,\x \right) = \frac{1}{2} \| \y-\B \x \|^2_2 + \lambda \| \x \|_1
\end{eqnarray}
where $\lambda$ is a positive scalar that controls the sparsity rate. In equation (\ref{eq:energy2}), the first term is smooth but the second one is not, and both are convex. Such least-square optimization problems with $\ell_1$-norm regularization are called \textit{Lasso}.

The set of feature vectors and the corresponding set of input vectors are stacked to form the matrices $\X=[\x_1, \cdots, \x_p]$ and $\Y=[\y_1, \cdots, \y_p]$, respectively. The energy function of (\ref{eq:energy2}) is optimized in two steps \cite{Mairal2012}:
\begin{enumerate}
\item In the \textit{learning phase}, feature vectors $\X$ corresponding to all input vectors $\Y$ are assumed to be fixed and the energy function in (\ref{eq:energy}) is minimized w.r.t. the dictionary matrix: 
    \begin{eqnarray} \label{eq:learn_phase}
    {\underset {\B}{\min}} & & \;
    \sum_{k=1}^p \frac{1}{2} \| \y_k-\B \x_k \|^2_2
    \\
    \textrm{subject} \; \ \textrm{to} &:& \; \| \b_j\|^2_2 \le c, \;\; j=1, \cdots, m \nonumber
    \end{eqnarray}
where $c$ is a constant, whose value is usually chosen to be 1, and $\b_j$ denotes a dictionary basis vector. Hence, in the learning phase, we are dealing with a quadratic programming problem, which can be solved by Lagrange \cite{Lee2006} or online dictionary learning \cite{Mairal2009} methods.
\item In the \textit{inference phase}, the dictionary matrix is assumed to be fixed and the energy function in (\ref{eq:energy2}) is minimized w.r.t. the feature vectors $\X$ for all input vectors $\Y$:
    \begin{eqnarray} \label{eq:inference_phase}
    {\underset {\X}{\min}} & & \;
    \sum_{k=1}^p \frac{1}{2} \| \y_k-\B \x_k \|^2_2 + \lambda \| \x_k \|_1
    \end{eqnarray}
    This phase needs more attention because of its non-smooth sparsity term.
\end{enumerate}
A solution for (\ref{eq:energy2}) is obtained by iteratively solving the above two subproblems \cite{Komodakis2015}.

Modified versions of sparse coding have been proposed in the literature in order to make sparse coding applicable to time-varying signals. The proposed approaches have benefitted from Kalman filtering \cite{Vaswani2008}, hierarchical Bayesian structures \cite{Karseras2013}, dynamic programming using homotopy \cite{Charles2011}, and reweighted $\ell_1$ dynamic filtering \cite{Charles2013}. In \cite{Chalasani2013} and \cite{Chalasani2014}, a dynamic version of sparse coding was proposed by adding another term to the objective function of (\ref{eq:energy2}). The new term is based on a state-space model for time-varying input signals that captures the temporal evolution of their features. In this version of sparse coding, a feature vector, $\x$, is viewed as a state vector, whose temporal evolution is governed by a linear state equation built on the first-order Markovian assumption. Hence, the optimization problem of (\ref{eq:inference_phase}) is modified as follows:
    \begin{eqnarray} \label{eq:inference_phase2}
    {\underset {\X}{\min}}  \;
    \sum_{k=1}^p \frac{1}{2} \| \y_k-\B \x_k \|^2_2 + \gamma \| \x_k-\A \x_{k-1} \|_1 + \lambda \| \x_k \|_1
    \end{eqnarray}
where $\A \in \mathbb{R}^{n \times n}$ is the state transition matrix. 

Sparse coding can be used to build encoders in deep learning structures for extracting invariant features from input images \cite{Gregor2010}, and dynamic sparse coding may be used for building encoders in deep predictive coding networks for feature extraction from video streams \cite{Chalasani2013, Principe2014}. In~\cite{Tao2017}, a semantic learning system was designed for tagging mobile images. A method based on sparse coding was proposed for speech unit classification in~\cite{Sharma2018}. Sparse coding was used to design classifier ensembles in~\cite{Tuysuzoglu2018} by random subspace dictionary learning (RDL), and bagging dictionary learning (BDL) algorithms were examined by learning ensembles of dictionaries through feature/instance subspaces. In~\cite{TorresBoza2018}, sparse coding was deployed for feature representation in speech emotion recognition. Facial motion recognition is another application of sparse coding \cite{Sunitha2017}. In \cite{Hu2018} and \cite{Murray2018}, sparse coding was used for designing deep neural network architectures. In \cite{Oguslu2018} and \cite{Qi2018} sparse coding was used for object detection and tracking.

In deep learning-based feature extraction, especially for real-time applications, it is critical to keep the number of model parameters at a minimal level while satisfying the desired performance criteria. Therefore, the advantage of using systematic approaches that avoid unnecessary parameters is two fold; they reduce both the computational burden and the required memory for saving the trained model~\cite{Deng2013, Squeezdet}. Following this line of thinking, metalearning can play a key role in developing such systematic approaches \cite{Clune2020}. Dimension of the feature vector is usually selected by the designer. It is usually chosen by trial and error, which would be challenging in occasions, where the number of input samples is very large, the input sample space is an uncountable set, or the input samples have high entropy, which in turn, may lead to large reconstruction errors \cite{cognitivecontrol2012, fatemi2017observability}. Although over-completeness provides robustness, computational burden increases with dimension of the feature vector. Therefore, there must be a trade-off between robustness and computational complexity.

In this paper, a method is proposed for changing the dimension of feature vectors on the fly based on reconstruction error. For instance, when sparse coding is used to build encoders in a deep learning structure, if the reconstruction error increases, the proposed method will augment the feature vector and increase its dimension on the fly instead of terminating the learning procedure and rerunning the program with a new dimension. A controller is devised based on variable structure learning automata, which learns to increase the feature vector dimension regarding the reconstruction error threshold in the reinforcement leaning framework. The proposed automata aims at keeping the reconstruction error below a threshold as well as increasing the dimension in a way to have the minimum possible number of bases. In this work, FISTA is deployed as the optimizer during the inference phase, and online dictionary learning (ODL) is used to optimize the dictionary in the leaning phase.

The rest of the paper is organized as follows. The proposed algorithm for increasing the feature vector dimension in a real-time manner is presented in Section~2 along with theoretical discussions. The proposed variable-structure automata is presented in section~3. Computer experiments are provided in Section~4, and the paper concludes in the final section.

\section{Sparse Coding with Variable-Dimension Feature Vectors}  
\label{method}

In online applications with large datasets such as object recognition in video streams \cite{Zhao2011}, deep predictive coding networks \cite{Chalasani2013}, or any other application, for which obtaining an estimate of the proper number of dictionary elements would be hard in the beginning, a mechanism is needed to increase dimension of the feature vectors and accordingly size of the dictionary matrix on the fly, when the reconstruction error increases beyond an acceptable level. Here, we focus on sparse coding with online learning, when each sample is processed separately. If reconstruction error increases, new basis vectors will be added to the dictionary and the feature vectors will be augmented with the coefficients associated with those new bases. 
If new basis vectors are added to the dictionary matrix and accordingly dimension of the feature vectors increases, it will be necessary to re-estimate the feature vectors associated with the previous input samples. This process calls for rerunning the program, which is time consuming. Hence, we need to look for a more efficient algorithm that can handle dimension variability. 

To address this issue, a method is proposed that does not require re-estimation of feature vectors associated with the previous input samples. Assume that in the inference phase, when the $p$th sample arrives, $\ell \le n$ new basis vectors will be needed. Then, dimension of the feature vectors must be increased by $\ell$. Before extending the dictionary matrix and feature vectors, $\X$  is an $n \times p$ matrix, whose columns are feature vectors associated with sample vectors 1 to $p$. After adding $\ell$ new basis vectors, $\X$ will be an $(n+\ell) \times p$ matrix. Instead of recalculating the feature vectors, the following rule is used to assign values to extra components of the feature vectors (i.e., elements $n+1$ to $n+\ell$):
    \begin{eqnarray} \label{eq:Xupdate}
    \X(n+1:n+\ell , 1:p) = 0 
    \end{eqnarray}
In other words, all feature vectors corresponding to input vectors prior to the $p$th sample are augmented with zero elements. Since new added bases have not been considered when previous input samples were handled, new feature vectors are initialized by random values. Hence, random values are assigned to dictionary bases related to extra feature dimensions as follows:
    \begin{eqnarray} \label{eq:Xupdate2}
    \B(:, n+1:n+\ell) = random \; values 
    \end{eqnarray}
After adding $\ell$ new bases, the augmented matrices will have the following dimensions: $\X \in \mathbb{R}^{(n+\ell) \times p}$ and $\B \in \mathbb{R}^{m \times (n+\ell)}$. If we consider a reconstruction capacity for dictionary bases, whenever the average reconstruction error of inputs increases, it means that the majority of bases are involved in reconstruction of previous inputs. With adding new bases we can hope to improve the reconstruction capacity for new inputs. Next section covers the reinforcement learning-based controller for dimension change.

\section{The Automata-based Controller} 
\label{controller}

As mentioned in the previous section, the reconstruction error can be decreased by increasing the feature vector dimension. Therefore, a controller must be designed to decide when and how much this dimension should be increased. Variable-structure learning automata would be a valid candidate for designing such a controller. The proposed controller deploys reinforcement learning to adjust the reconstruction error in a dynamic manner by choosing an appropriate dimension for feature vectors. In other words, the controller aims at keeping the reconstruction error below a predefined threshold by minimal increment of the dimension. 

\begin{figure*}[]
\centering
\includegraphics[width=0.75\linewidth]{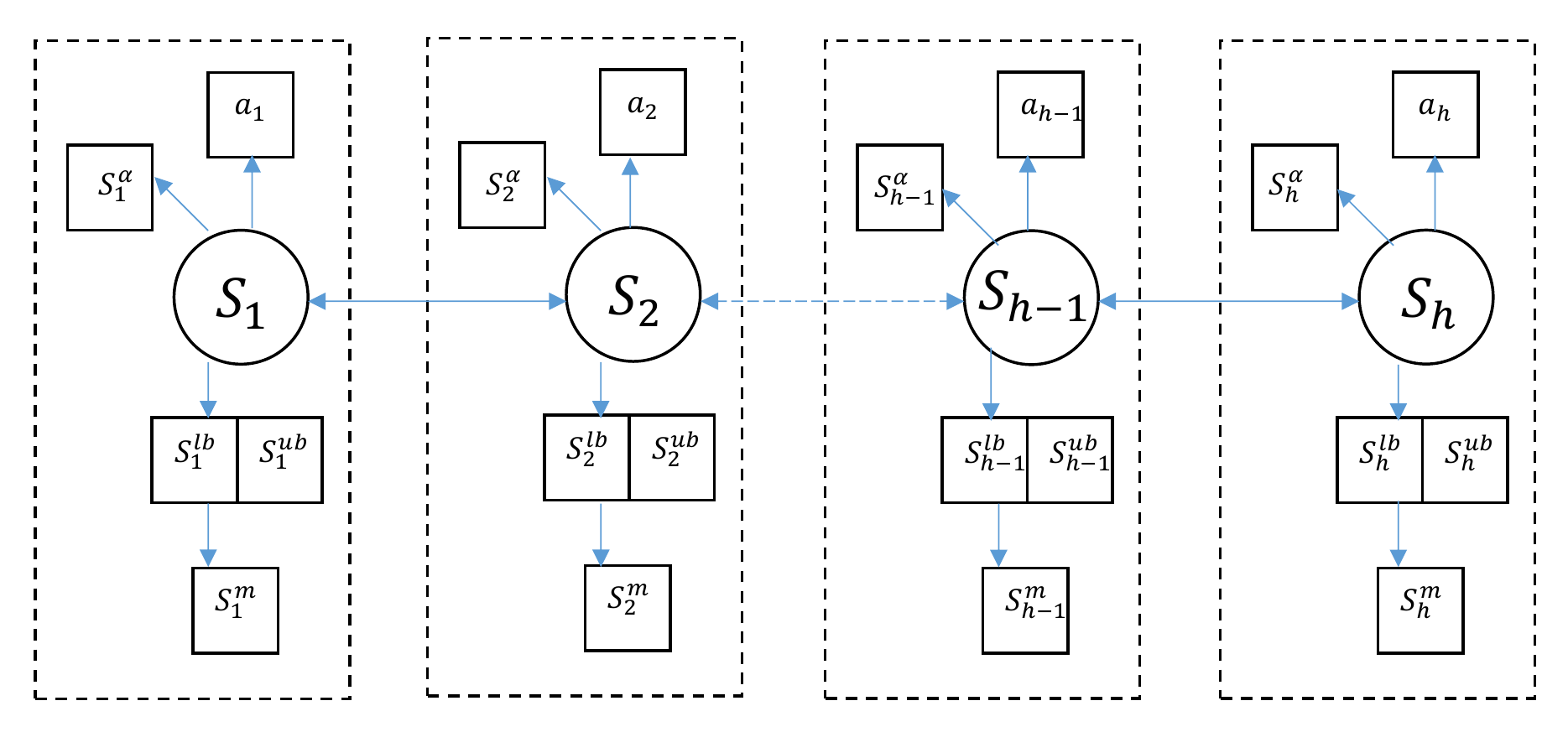}
\caption{Architecture of the proposed automata, where $s_i$ and $a_i$ refer to states and actions, respectively, $s_i^{lb}$ and $s_i^{ub}$denote the lower and the upper bounds on the reconstruction error, and $s_i^{m}$ refers to memory.}
\label{fig:1}
\end{figure*}

In the reinforcement learning framework, an agent (decision maker or controller in this context) takes actions in an environment in a way to maximize the expected collected reward (return) over the desired control horizon. The agent tries to find an optimal policy via interactions with its environment. Markov decision process (MDP) provides the mathematical framework for formulating a reinforcement learning problem. The corresponding MDP is defined by a set of states, a set of actions, a state transition probability, a reward function, and a discount factor. Here, the reinforcement learning algorithm is built around  variable-structure learning automata with memory. Figure \ref{fig:1} shows the architecture of the proposed automata, which is called sparse coding automata (SCA). The SCA is mathematically defined as $automata = \{S, A, E, f, g\}$, where $S=\{s_1,s_2,\cdots,s_h\}$ is the set of automata states, 	$A=\{a_1,a_2,\cdots,a_r\}$ is the set of automata actions, $E=\{e_1,e_2,\cdots,e_q\}$ is the set of inputs, $f: e \rightarrow s$ is the transition function that determines one of the $h$ states according to the $k$th input, and $g: s \rightarrow a$ is the output function that opts for one of the $r$ possible actions based on the state, which has emerged from the transition function. 

The proposed SCA increases the dimension of the feature vectors in online sparse coding, when the reconstruction error violates a predefined constraint. SCA aims at finding the minimum dimension for feature vectors that keeps the reconstruction error below the desired threshold. In this way, all inputs up to the current one are reconstructed using the minimum possible feature dimension. The input to the automata is defined as the mean-squared reconstruction error $e_k$ calculated for all inputs up until to the $k$th one. In SCA, an interval is assigned to each state $s_i$. At each instant $k$, the transition function $f$ selects the state $s_i$ corresponding to the input $e_k$ (i.e., $f(e_k )=s_i$), if $s_i^{lb}<e_k<s_i^{ub}$, where $lb$ and $ub$ denote the lower and upper bounds of state $s_i $. In addition, there is a constant action $a_i$ associated with each state $s_i$, which increases the dimension of feature vectors by the amount of $\ell_i$. It is assumed that  $\ell_1 < \ell_2 < \cdots < \ell_h$. To be more precise, function $g$ selects a certain action $a_i$ at each state $s_i$ of the automata, which is in turn, selected by function $f$ based on the interval that includes the reconstruction error. Since both functions $f$ and $g$ are deterministic mappings, SCA is a deterministic automata. Algorithm~1 provides the pseudocode for sparse coding combined with the proposed reconstruction error control, which is based on variable structure automata. Algorithm~2 provides the pseudocode for changing the dimension of feature vectors, which is used as a subroutine in Algorithm~1.

\begin{algorithm} \label{algorithm1}
\SetAlgoLined
\textbf{Variables:}\\
$\Y=[\y_1, \cdots, \y_p] \in \mathbb{R}^{m \times p}$: matrix of all input vectors \\
$\X=[\x_1, \cdots, \x_p] \in \mathbb{R}^{n \times p}$: matrix of all feature (i.e. coefficient) vectors \\
$\B \in \mathbb{R}^{m \times n}$: dictionary matrix \\

\textbf{Initialization:} \\
$k=0$;
Initialize $\x_0$  and $\B_0$

\Repeat{the last frame is processed}{
$k\leftarrow k+1$\

\line(1,0){210}\\
\textbf{Calculate the feature vector $\x_k$ for the $k$th frame and the dictionary matrix $\B$:} \\
	\While{convergence condition has not been met}{
		Update $\x_k$ using FISTA \cite{Beck2009} or dynamic sparse coding \cite{Chalasani2013}\\
        Update $\B$ using online dictionary learning	\cite{Mairal2009}  
	}

\line(1,0){210}\\
\textbf{Change the dimension if it is necessary:} \\
\If{the reconstruction error is unacceptable}{
Run Algorithm 2
}

}

\caption{Sparse coding automata}
\end{algorithm}

\begin{algorithm} \label{algorithm2}
\SetAlgoLined

\textbf{Initialization:} \\
$s_i^{lb}=0$, $s_i^{ub}=\infty$, $s_i^m=0.5$, $s_i^a=\infty$; for $i=1 \cdots h$ \\
$0 < \sigma < 1$

\textbf{Inputs:} \\
$a_k$: action \\
$e_k$: mean-squared reconstruction error

\If{$e_k > threshold$}{

	\If{$e_k < s_i^a$}{
		$s_i^a = e_k$
}
	$penalty = \sigma (threshold-e_k)$ \\
	$s_m^i = penalty$ \\
	\If{$s_i^m \ge 1$}{
		$s_i^{ub}= s_i^a$ \\
		$s_{i+1}^{lb}=s_i^a$ \\
        $s_i^m=0.5$
}
}
\caption{Variable-structure automata algorithm}
\end{algorithm}

In the learning algorithm, structure of function $f$ is updated dynamically due to the reconstruction error changes. In fact, the learning phase is activated when the reconstruction error violates the threshold. As mentioned before, an interval $(s_i^{lb},s_i^{ub})$ is assigned to each state $s_i$. In fact, the automata learns to take action $a_i$, which corresponds to state $s_i$, when the reconstruction error falls in this interval. At the beginning, all states have the same interval but the automata will select the state with the minimum action value. The automata will be penalized in a certain state, if the corresponding action cannot reduce the reconstruction error below the predefined threshold. According to the penalty, the upper bound of the state changes to achieve the minimum reconstruction error achievable by the automata $s_i^a$. Moreover, the lower bound of the next state $s_{i+1}^{lb}$ is replaced by this value. Here, $\sigma$ is a parameter that represents the memory size $s_i^m$. This memory allows for providing a degree of smoothness in changing the state, when the automata tries to compensate for the penalty. SCA can be viewed as a non-parametric model, which seeks to model the implicit dynamics in data without making any prior assumption about the model. In other words, SCA does not confine the solution to a particular function.

\section{Experimets} 
\label{experiment}

\begin{figure*}[]
\centering
\includegraphics[width=1\linewidth]{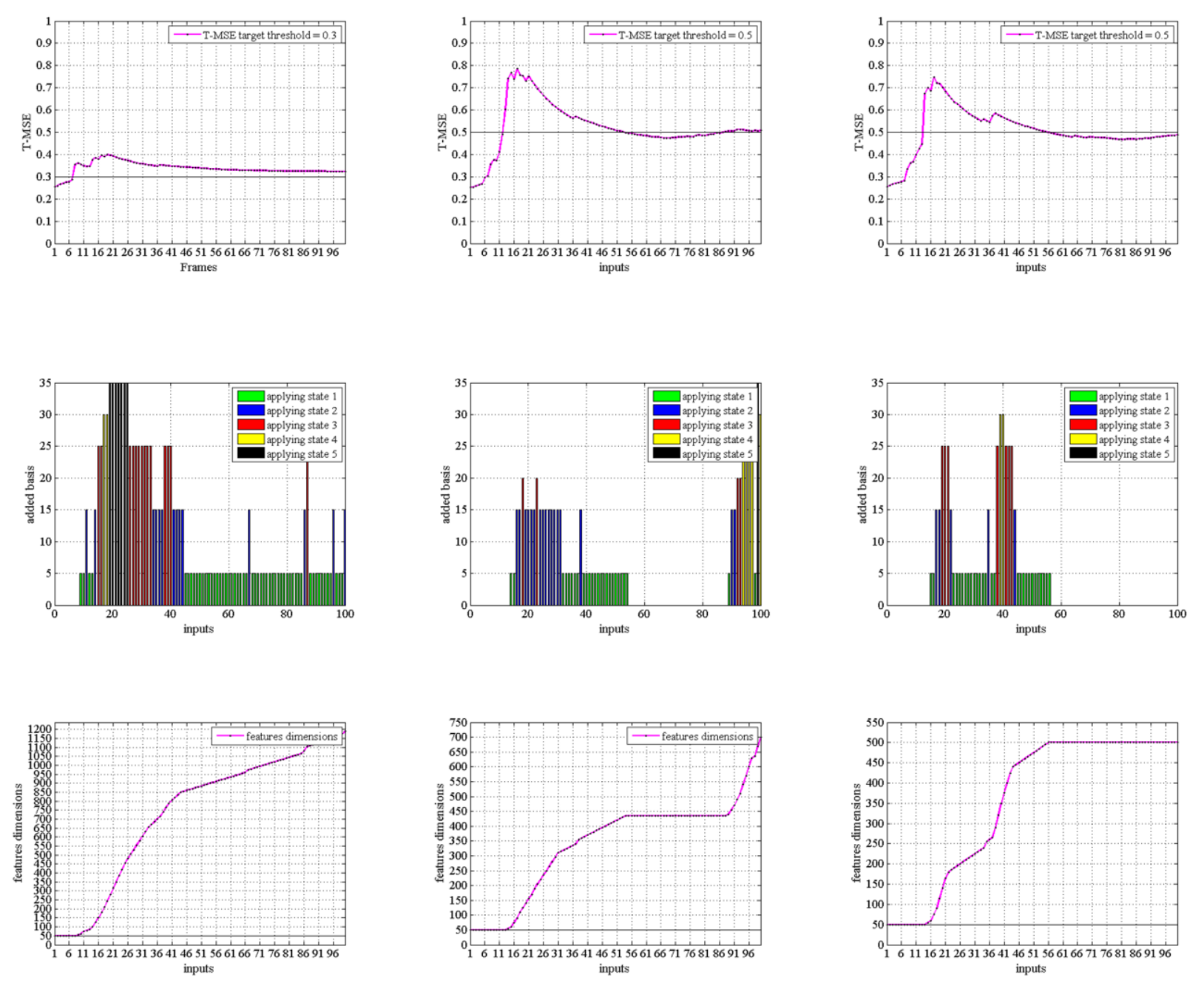}
\caption{For the three studied configurations, the first row depicts evolution of T-MSE as more inputs are presented to the sparse coding automata algorithm. The second row shows the visited states and the applied corresponding actions. The third row demonstrates the evolution of feature vector dimension.}
\label{fig:4}
\end{figure*}

To evaluate the performance of the proposed SCA, 100 images from the Caltech 101 dataset were selected. Each image was resized to a $40 \times 40$ matrix. Each resized image was divided to four $20 \times 20$ patches. Then, each patch was converted to a vector $y \in  \mathbb{R}^{400}$. All of the patches were converted to gray scale. These 400 patches were used as inputs to the proposed SCA algorithm to evaluate a variety of configurations. To illustrate the capacity of the dictionary of bases, traditional sparse coding with a constant feature dimension was tried as well to reconstruct inputs. Both online dictionary learning and FISTA were used \cite{Mairal2009, Beck2009}. The total mean-squared error (T-MSE) is expected to increase when more inputs are reconstructed. To evaluate the performance of SCA, three different system configurations were tested: 
\begin{itemize}
\item The automata has five state $h=5$, five actions $A=\{5,15,20,30,35\}$, the reconstruction threshold is equal to 0.5, and the system started with feature vector dimension of 50. 
\item The automata has five state $h=5$, five actions $A=\{5,15,25,30,35\}$, the reconstruction threshold is equal to 0.5, and the system started with feature vector dimension of 50. 
\item The automata has five state $h=5$, five actions $A=\{5,15,25,30,35\}$, the reconstruction threshold is equal to 0.3, and the system started with feature vector dimension of 50. 
\end{itemize}
The memory parameter was assumed to be $s_i^m=0.5$ for all states. Subplots in the first row of Figure \ref{fig:4} illustrate the T-MSE for these three scenarios. The SCA was applied when the T-MSE reached a predefined threshold. After a period of learning, SCA was able to control the T-MSE by taking appropriate actions as shown in subplots in the second row of Figure \ref{fig:4}. The third row of Figure \ref{fig:4} shows the evolution of feature vector dimension over time as more inputs are presented to the SCA algorithm. The system executed Algorithm~1 after processing of four patches of input images. Using memory paves the way for reducing the effect of outliers on learning the lower and upper bounds of the states, and achieving a degree of smoothness in the learning process. As shown in Figure \ref{fig:4}, the automata can successfully control the reconstruction error. Table~1 summarizes the learned structures for function $f$ (state interval) for these configurations.

\begin{table*}[ht] \label{tbl:1}
\caption{Results of the inner structure of function $f$ in three configurations of the automata}
\vspace{0.2cm} \centering
\begin{tabular}{c c c c c c c c c c c}
\hline \hline  
 & $(s_1^{lb} , s_1^{ub})$ & $(s_2^{lb} , s_2^{ub})$ & $(s_3^{lb} , s_3^{ub})$ & $(s_4^{lb} , s_4^{ub})$ & $(s_5^{lb} , s_5^{ub})$ \\
[0.5ex]
\hline \hline \\
Configuration~1 &	(0 , 0.5090) & (0.5090	, 0.5582) & (0.5582 , 0.5644)	& (0.5644 , 0.5780) & (0.5780 , $\infty$)  \\
[1ex]

 \\
Configuration~2 &	(0 , 0.5047) & (0.5047	, 	0.5050) & (0.5050 , 0.5052)	& (0.5052 , 0.5066) & (0.5066 , $\infty$)  \\
[1ex] 

 \\
Configuration~3 &	(0 , 0.3239) & (0.3239	, 0.3242) & (0.3242 , 0.3253)	& (0.3253 , 0.3787) & (0.3787 , $\infty$)  \\
[1ex]

\hline
\end{tabular}
\end{table*}

\begin{figure*}[]
\centering
\includegraphics[width=1\linewidth]{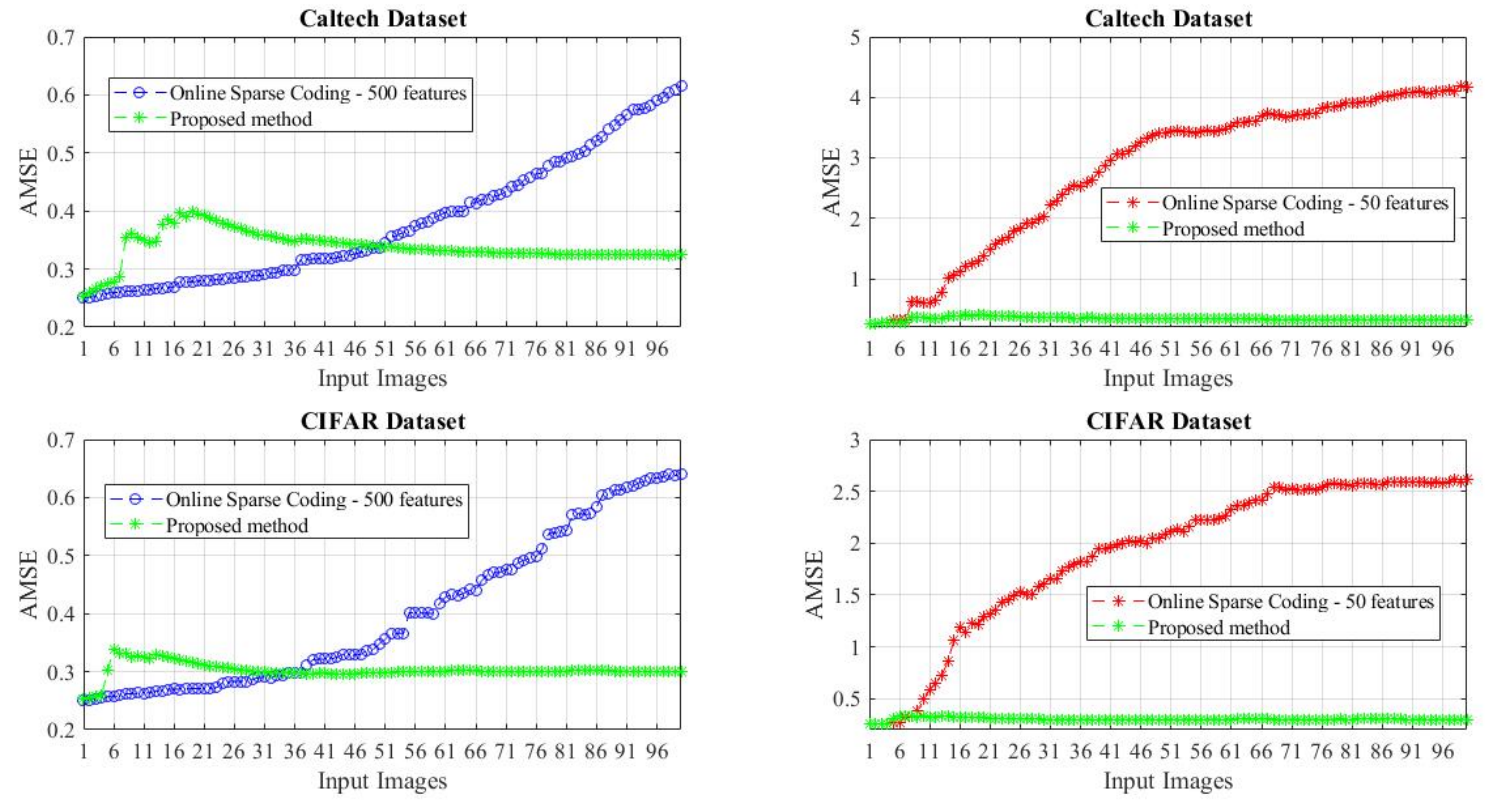}
\caption{The first row shows the results of the SCA and sparse coding with two configurations on Caltech dataset. The second row provides the same information on CIFAR-10 dataset.}
\label{fig:5}
\end{figure*}

In addition, to evaluate the performance of the SCA in encoding different datasets in comparison with the original version of the sparse coding, additional experiments were performed on Caltech 101 and CIFAR-10 datasets as benchmarks. In this regard, two sets of data were generated, which included 100 images from each one of these two datasets. Experiments were performed with the three mentioned configurations for SCA and sparse coding with the aim of encoding the chosen dataset. Again, each image was converted to gray scale and resized to a $40 \times 40$ matrix and divided to four $20 \times 20$ patches. Results of the first configuration as the architecture of automata are reported here on the benchmark datasets. In the experiment, SCA was supposed to control the T-MSE to remain about $0.3$. Sparse coding was initialized with $50$ and $500$ feature elements so that both over-complete and under-complete dictionary can be taken into account in the comparision. Figure \ref{fig:5} depicts the results on the benchmark datasets. Subplots in the first row illustrate the performance of models for Caltech dataset, while the second row provides similar information for CIFAR dataset. Subplots in the first column show that the SCA was able to keep the T-MSE around the $0.3$, while the sparse coding with under-complete dictionary cannot represent the data with a fairly constant precision and the T-MSE increases over time. A similar increasing pattern in the T-MSE is observed for the sparse coding with over-complete dictionary but with a lower error rate. As it can be seen, the SCA was able to control the error of reconstructions for both benchmarks.  

\begin{figure*}[]
\centering
\includegraphics[width=1\linewidth]{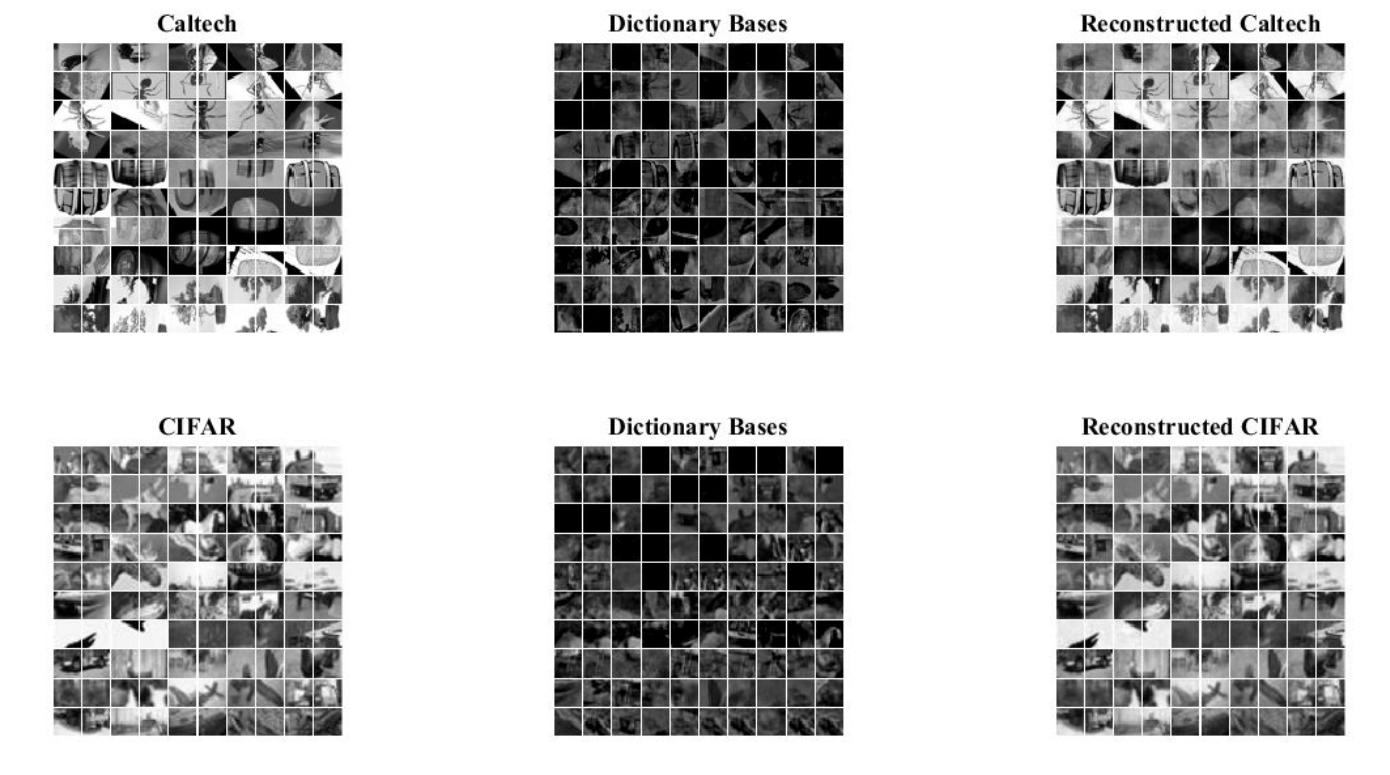}
\caption{Reconstructed image patches (left), bases of the learned dictionary (middle), and the original input image patches (right).}
\label{fig:6}
\end{figure*}

Figure \ref{fig:6} shows the original input image patches, dictionary bases, and the reconstructed input patches for the first configuration on both datasets. As shown, SCA can successfully control the reconstruction error by choosing an appropriate minimal dimension for feature vectors. 

\section{Concluding Remarks} 
\label{conclude}

This paper proposed a simple method to expand the dictionary matrix and the corresponding feature vectors in encoders, which are built around sparse coding. The proposed algorithm allows for changing the dimension of the feature vectors on the fly when the reconstruction error is unacceptable without re-estimating the feature vectors associated with previous samples. This method provides an over-complete representation by automatically selecting an appropriate number of basis vectors. Thus, it will not be necessary to initially choose a large number of basis vectors to achieve over-completeness. Moreover, trial and error for choosing a proper dimension can be avoided. Hence, the proposed algorithm provides a systematic way for setting up the feature vector dimension in order to reach the best reconstruction error with minimal computational cost. The memory size required for saving the trained model will be minimal as well.

\bibliographystyle{IEEEtran}
\bibliography{IEEEabrv,Main}

\end{document}